\documentclass[preprint,12pt]{elsarticle}

\usepackage{color,array}
\usepackage{anysize}
\usepackage{subfigure}
\usepackage{graphicx}
\usepackage{float}
\usepackage{multirow}
\usepackage{tabularx}
\usepackage{booktabs} 
\usepackage{caption}
\usepackage{lineno}
\usepackage{xspace}
\usepackage{amsmath}

\usepackage[]{graphicx}
\usepackage{subfigure}
\usepackage{setspace}

\usepackage{comment}

\begin{document}

\begin{frontmatter}

\title{Denoising of 3D magnetic resonance images with multi-channel residual learning of convolutional neural network}

\author[address1,address2]{Dongsheng Jiang}
\author[address3]{Weiqiang Dou\fnref{myfootnote}}
\fntext[myfootnote]{Dongsheng Jiang and Weiqiang Dou have equal contributions on this work.}
\author[address4]{Luc Vosters}
\author[address5,address6]{Xiayu Xu}
\author[address4]{Yue Sun}
\author[address7,address8]{Tao Tan\corref{mycorrespondingauthor}}
\cortext[mycorrespondingauthor]{Corresponding author; tao.tan911@gmail.com}

\address[address1]{Digital Medical Research Center, School of Basic Medical Science, Fudan University, Shanghai, China}
\address[address2]{Shanghai Key Laboratory of Medical Image Computing and Computer-Assisted Intervention, Shanghai, China}
\address[address3]{Department of Radiology and Nuclear Medicine, Radboud University Medical Center,  Nijmegen, the Netherlands}
\address[address4]{Department of Electrical Engineering, Eindhoven University of Technology, Eindhoven, the Netherlands}
\address[address5]{Bioinspired Engineering and Biomechanics Center (BEBC), Xi’an Jiaotong University, Xi’an 710049, China}
\address[address6]{The Key Laboratory of Biomedical Information Engineering of Ministry of Education, School of Life Science and Technology, Xi’an Jiaotong Universty, Xi’an 710049, P.R. China}
\address[address7]{Department of Biomedical Engineering, Eindhoven University of Technology, Eindhoven, the Netherlands}
\address[address8]{ScreenPoint Medical, Nijmegen, the Netherlands}

\begin{abstract}

The denoising of magnetic resonance (MR) images is a task of great importance for improving the acquired image quality. Many methods have been proposed in the literature to retrieve noise free images with good performances. Howerever, the state-of-the-art denoising methods, all needs a time-consuming optimization processes and their performance strongly depend on the estimated noise level parameter. 
Within this manuscript we propose the idea of denoising MRI Rician noise using a convolutional neural network. The advantage of the proposed methodology is that the learning based model can be directly used in the denosing process without optimization and even without the noise level parameter. Specifically, a ten convolutional layers neural network combined with residual learning and multi-channel strategy was proposed. Two training ways: training on a specific noise level and training on a general level were conducted to demonstrate the capability of our methods.
Experimental results over synthetic and real 3D MR data demonstrate our proposed network can achieve superior performance compared with other methods in term of both of the peak signal to noise ratio and the global of structure similarity index. Without noise level parameter, our general noise-applicable model is also better than the other compared methods in two datasets. Furthermore, our training model show good general applicability. 
 
\end{abstract}

\begin{keyword}
MRI, denoising, CNN, Rician noise, deep learning
\end{keyword}

\end{frontmatter}


\section{Introduction}
\label{sect:intro}  

Magnetic resonance imaging (MRI), as an attractive non-invasive medical imaging technique, plays an important role on the diagnosis of pathological and physiological alterations in living tissues and organs of human body, as it can provide high resolution three dimensional (3D) images with anatomical and functional contrast due to different MR measurable tissue parameters. The quality of MR images can, however, easily be degraded by random noise generated during image acquisition. Increased noise level can dramatically affect the accuracy of diagnoses and also the reliability of quantitative image processing including segmentation, registration and classification. To avoid these problems, it is therefore essential to remove the noises of MR images (image denoising) before implementing further image processing.

In the past years a number of image denoising methods have been reported. Martin-Fernandez and Villullas \cite{Martin-Fernandez2015}  applied a new method with shrinkage of wavelet coefficients based on the conditioned probability of being noise. The involved the parameters are calculated by means of the expectation maximization method. In addition, Change et al.\cite{Chang2015} took full use of the block representation advantage of nonlocal means 3D method (NLM3D) to restore the noisy slice from different neighboring slices and employed a post processing step to remove noise further while preserving the structural information of 3D MRI. Zhang et al. \cite{Zhang2015} investigated the application and improvement of  higher-order singular value decomposition (HOSVD) to denoise MR volume data and achieved comparable performance to that of block-matching 4D method (BM4D). Manj\'{o}n further et al\cite{Manjon2015} proposed a novel method for MRI denoising that exploits both the sparseness and self-similarity properties of MR images with the state-of-art performance. Baselice et al.\cite{Baselice2017} exploited Markov random fields to achieve the combination of details preservation and noise reduction without any supervision. Bhujle and Chaudhuri \cite{Bhujle2013} carried out nonlocal means denoising on squared magnitude images and thus compensated the introduced bias. Chang and Change \cite{Chang2015a} applied an artificial neural network associated with image texture feature analysis to establish a predictable parameter model which is able to automate the denoising procedure.  Golshan and Hasanzadeh \cite{Golshan2015} proposed a new filtering method based on the linear minimum mean square error (LMMSE) estimation, which employs the self-similarity property of MR data to restore the noise-less signal. Varadarajan and Haldar \cite{Varadarajan2015} described a novel majorize-minimize framework that allows penalized maximum likelihood estimates obtained by solving a series of much simpler regularized least-squares surrogate problems.  While robust denoising performances are usually able to be achieved, most of these methods, however, include complex optimization processes for images \cite{Zhang2017a} and therefore are time-consuming. Furthermore, these methods generally involve the optimization of a non-convex cost function and requires a manual parameter selection, meaning that the above methods usually need a procedure to estimate the noise level parameter.

To avoid optimization in test phase and directly apply discriminant learning, in this study we extended feed-forward denoising conventional neural networks (DnCNNs) which was originally proposed by by Zhang et al.\cite{Zhang2017a} to restore the noise-free MR images from the noisy ones. To adapt to 3D volume, we applied  a multi-channel approach for learning that we hypothesize will a faster and stabler training as well as a more robust denoising performance.  To validate these, four other state of art denoising methods were also employed and compared with our proposed method. To our best knoweldge, this paper shows the first deep learning based method to denoise MR images with Rician noise.

\section{Method}

\subsection{MCDnCNN}

The DnCNNs  proposed by Zhang et al. \cite{Zhang2017a} is developed from a VGG network to make it suitable for image denoising. To apply this model in 3D data, we used the multi-channel version of DnCNNs (MCDnCNN). The goal of this network is to compute the noise-free image. He et al. \cite{He} pointed out that the network must preserve all input details since the image is discarded and the output is generated from the learned features alone. With many weight layers, this becomes an end-to-end relation requiring very long-term memory. For this reason, the vanishing/exploding gradients problem can be critical. It was shown that \cite{He} the residual networks are easier to be optimized, and can gain accuracy from considerably increased depth. Therefore rather than directly outputing the noise-free image, the network is designed to predict the noise $v_i$ at each pixel of the original image $y_i$.  Therefore, the proposed DnCNN implicitly removes the latent clean image with the operations in the hidden layers. The noise free intensity can be computed as $y_i - v_i$. 

\begin{figure}[t]
\centering
\quad
{
\includegraphics[width=0.7\textheight]{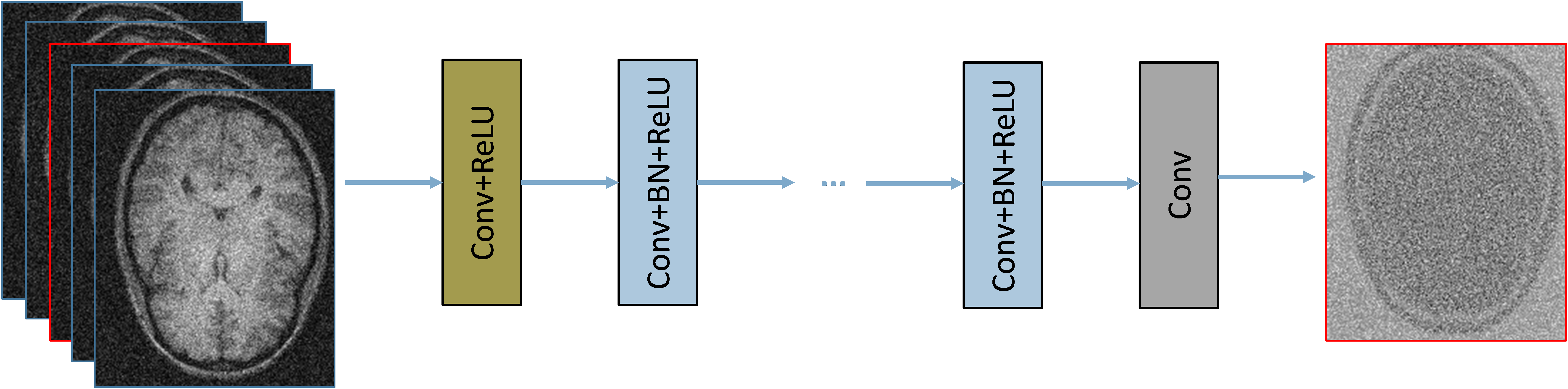}
} 
\caption{\label{fig:network} The layout of our proposed MCDnCNN network }
\end{figure}

This network, as illustrated in Fig \ref{fig:network}, consists  of one input layer of convolution with rectified linear unit (ReLU),  eight layers  of convolution  with  batch  normalization and RelU and one output layer of convolution.  The output layer generates an image with the same size as the  input. This network does not include any max-pooling layer as the output should have the same size as the input. For  the  first layer,  we use 64 kernels of size 3x3x$c$. For grey level 2D image, the $c$ can be set to 1. In our application, we are dealing with MR volume. One option is to process the volume slice by slice and $c$ is 1 accordingly. In this study, to make use of context information across neighboring slices of a particular slice $s$, we set the size of the input to be $X$ by $Y$ by 5. $X$ and $Y$ are the size of each 2D slice and the number of neighboring slices including slice $s$ is 5. The kernels of following layers including the last layer are also set to have a size of 3x3x64.  Two key features\cite{Zhang2017a} of this network, residual learning formulation and batch normalization, are incorporated to speed up training as well as to boost the denoising performance.

To optimize parameters of this proposed residual network, the averaged mean square error between the desired residual image and the estimated ones from the noisy input is calculated as the loss function of the networks:

\begin{equation}
  l(\theta) =  \frac{1}{2N}\sum_{i=1}^{N} ||R(y_i;\theta)-(y_i-x_i)||^2
\end{equation}

After computing the noise for every slice across the volume we can restore the original noise free volume.

So far, most existing state-of-the-art denoising methods depends on a fixed noise level. When applied to a MRI volume with unknown level of Rician noise, the common way is to first estimate the noise level parameter, and then denoise using the parameter. This makes the denoising results affected by the accuracy of noise estimation. To demonstrate the capability of DnCNN in general image denoising, we conducted two different training approaches. We first trained the DnCNN model on a specific noise level (MCDnCNNs) to evaluate the superior performance of our proposed deep learning approach compared with the-state-of-the-art Rician noise denoising approaches. Second, we extended our DnCNN model to a general model (MCDnCNNg) for Rician denosing with unknown noise level.

\subsection{ Comparison with other methods}

To evaluate the denoising performance of our method on 3D MR images, four well-established denoising methods have also been tested and compared in this study. 

\subsubsection{optimized block-wise non-local means (NLM) denoising filter (Coupe)}

This method was developed by Coupe P et al. \cite{Coupe2008}, based on the classic NLM denoising filter. Four new features, including a fully-automated tuning of the smoothing parameter, a selection of the most relevant voxels, a blockwise implementation and a parallelized computation have been combined into this filter, resulting in a better denoising performance with a significant decrease of computation time. 

\subsubsection{3D wavelet subbands mixing method (WSM) }

This method was an extension of the optimized block-wise NLM filter mentioned above. Coupe P et al \cite{Coupe2008a} proposed a mixing of a 3D sub-bands wavelet on this filter to further improve the performance of image noise removal, while keeping the computation time comparable.   

\subsubsection{3D oracle-based discrete cosine transform (DCT) filter (ODCT3D) }

The local DCT denoising filter was adapted\cite{Manjon2012a} to deal with Rician noise by using a pseudo-oracle principle. Compared with the two filters mentioned above, improved denoising results have been shown by using this filter with much shorter computation time (81s, 110s vs 10s).  

\subsubsection{3D prefiltered rotationally invariant nonlocal means filtering (PRI\_NLM3D) }

3D PRINLM filter is a new rotationally invariant version of the original NLM filter.  It was developed \cite{Manjon2012a} with the usage of the prefiltered image obtained by the above-mentioned DCT denoising. Compared to the other three filters, PRINLM filter was reported to have the best denoising performance with less than 1 minute processing time.

\subsection{ MR image acquisition and evaluation}

We have applied our training and test scheme on three subsets from two public datasets: IXI dataest and Brainweb datasets. One critical problem of deep learning approach is the weak general applicability. Networks trained on one dataset from a specific manufacture or setting may not perform well in a different dataset. To investigate the general applicability of our network, we collected different datasets in which one was used for training on one dataset while the other datasets were tested.

\subsubsection{IXI-Hammersmith dataset}
Hammersmith dataset is a subset of IXI dataset(http://brain-development.org/ixi-dataset/). The data was acquired in Hammersmith Hospital using a Philips 3T system (T1 parameters; Repetition Time = 9.60; Echo Time = 4.60; Number of Phase Encoding Steps = 208; Echo Train Length = 208; Reconstruction Diameter = 240.0; Acquisition Matrix = 208 x 208; Flip Angle = 8.0).

In total 30 sets of T1 weighted MR brain images from healthy subjects were randomly selected from IXI Database . 20 out of 30 images were used for training the neuralnetwork with the left 10 images for evaluating the described methods. 

For Rician denoising model training with either known or unknown noise level, we set the patch size as 60 x 60, and use sliding window strategy to obtain 150,000 patches to train the corresponding model. We use Adam for stochastic optimization and a mini-batch size of 64. We train 50 epochs for our two different models. The learning rate was decayed exponentially from 1e−1 to 1e−4 for the 50 epochs. MatConvNet package was used to train the proposed models.

\subsubsection{IXI-Guys dataset}
The second dataset is Guys dataset with 10 images acquired at Guy’s Hospital using a Philips 1.5T system(T1; Repetition time = 9.813; Echo time = 4.603; Number of Phase Encoding Steps = 192; Echo Train Length = 0; Reconstruction Diameter = 240; Flip Angle = 8). 

\subsubsection{Brainweb dataest}
The second dataset is Simulated Brain Database (http://brainweb.bic.mni.mcgill.ca/brainweb/). The SBD contains a set of realistic MRI data volumes produced by an MRI simulator，which were widely used to evalulate the performance of donoising approaches\cite{Feng2016,Zhang2014,AksamIftikhar2014,Coupe2008b}.

To evaluate the performance of the denoising from our proposed methods and other methods, we applied these methods on the data with different noise level from 1\% to 15\% with an increase of 2\%. We use two quantitative measures\cite{Zhang2015}. One is the peak signal to noise ratio (PSNR) which is defined as:

\begin{equation}
  PSNR =  20log_{10} \frac{255}{RMSE}
\end{equation}

\noindent where $RMSE$ is the root mean square error between denoised data and the noise-free data. The second measure is the global of structure similarity index (SSIM) which is the average of structure similarity index ($SSIM_{l}$) \cite{Manjon2012} defined as :

\begin{equation}
  SSIM_{l} =  \frac{(2\mu_{x}\mu_{\hat{x}}+c_{1} )(2\sigma_{x\hat{x}}+c_{2}) }{(\mu_{x}^2+\mu_{\hat{x}}^2+c1)(\sigma_{x}^2+\sigma_{\hat{x}}^2+c2) }
\end{equation}

\noindent where $\mu_{x}$ and $\mu_{\hat{x}}$ are the mean of the data $x$ and $\hat{x}$ and $c_{1}$ and $c_{2}$ are constants. $\sigma_{x}$ and $\sigma_{\hat{x}}$ are the variances and $\sigma_{x\hat{x}}$ is the covariance of $x$ and $\hat{x}$.  $SSIM_{l}$ has been computed with a 3x3x3  voxel kernel.

\section{Results}

\subsection{Results from IXI-Hammersmith dataset}

\begin{figure}[t]
\centering
\quad
\subfigure[]{
\includegraphics[width=0.35\textheight]{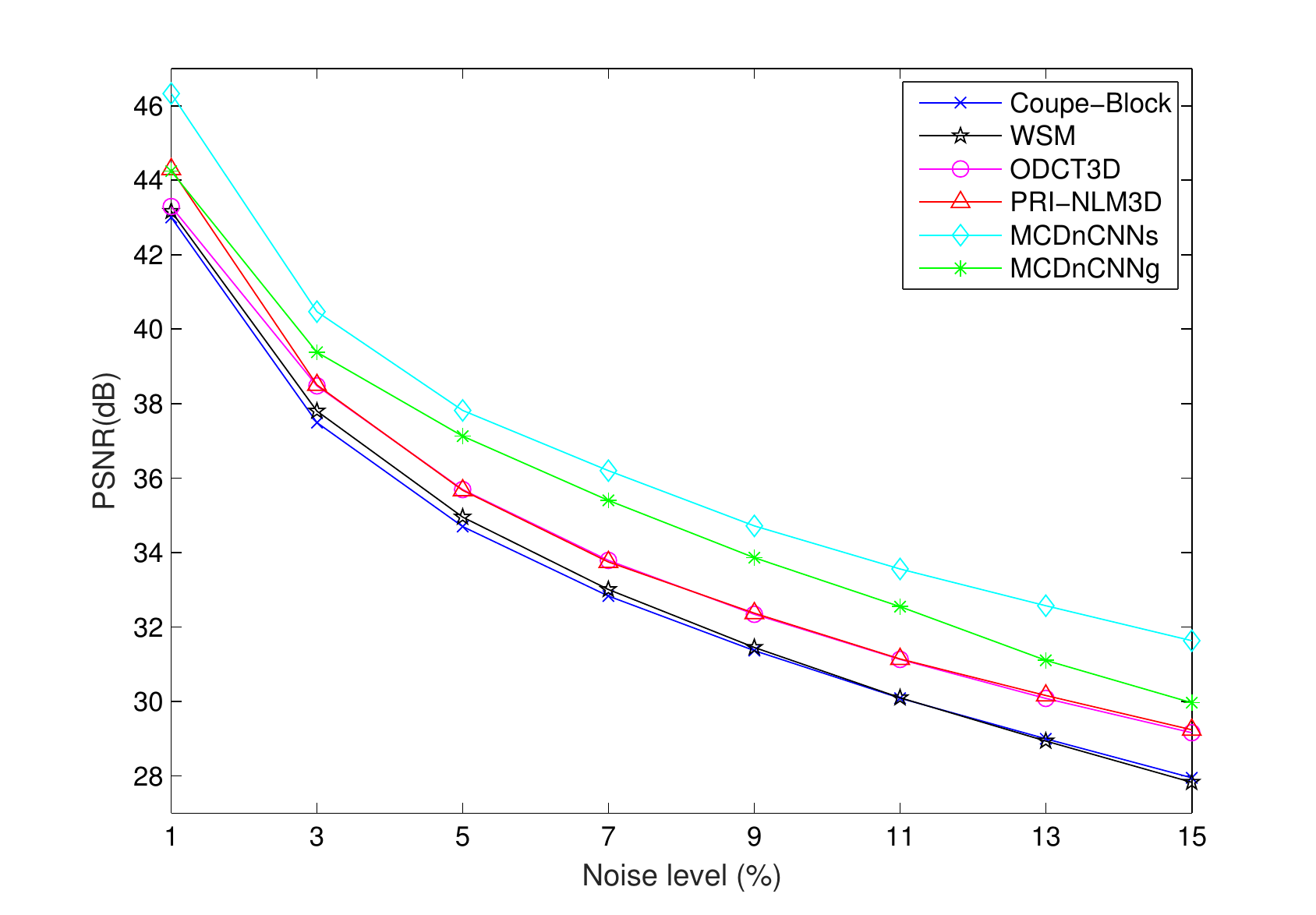}
}
\subfigure[]{
\includegraphics[width=0.35\textheight]{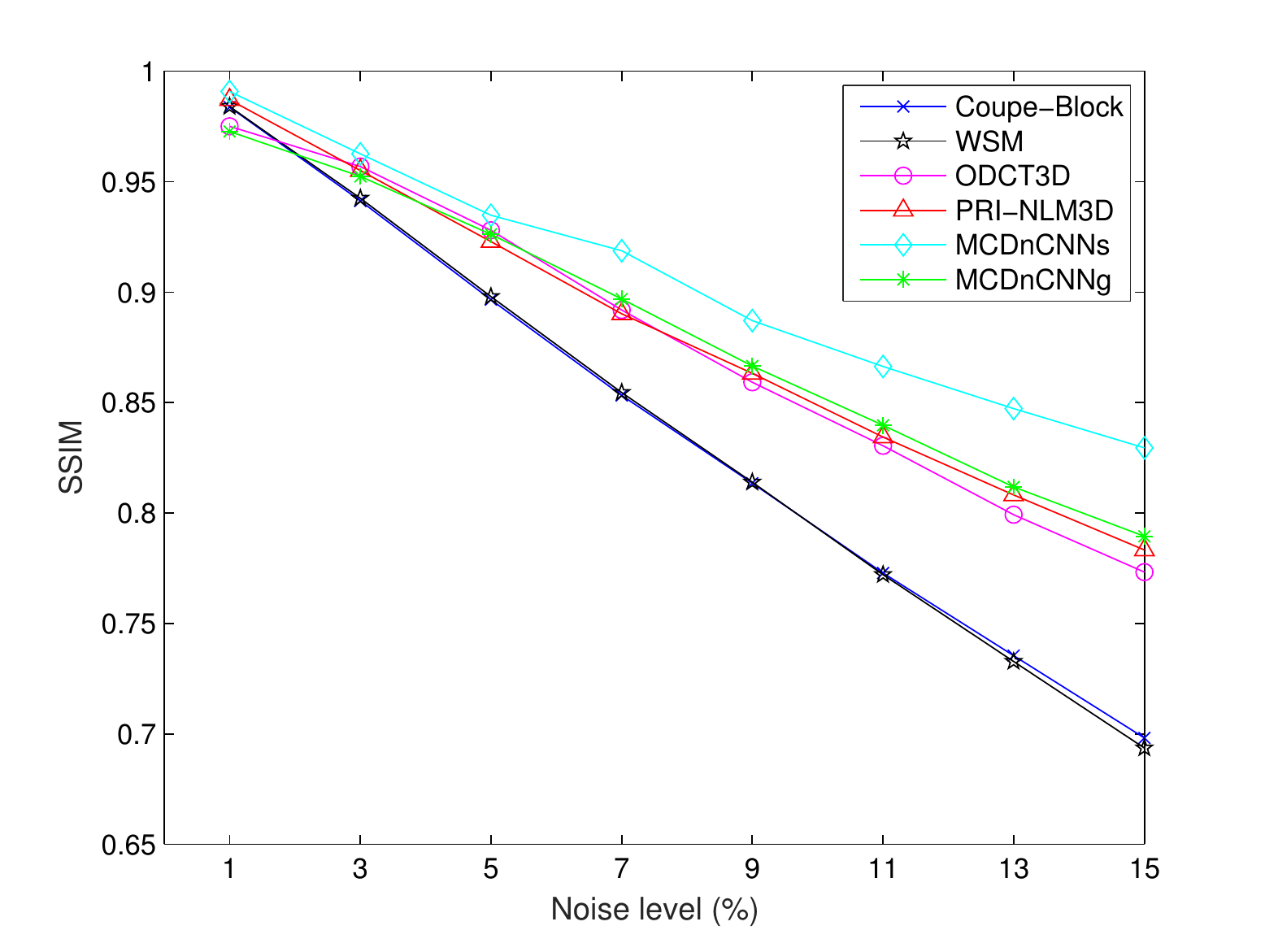}
} 
\caption{\label{fig:measures} De-nosie measures of PSRN (a) and SSIM (b) from different methods with different noise levels from IXI-Hammersmith dataset}
\end{figure}

Tables \ref{table:PSRN} and \ref{table:SSIM} and Figure \ref{fig:measures} summarize the PSNR and SSIM measures in IXI-Hammersmith dataset with different methods . The noise-specific model MCDnCNN (MCDnCNNs) significantly outperformed all other methods in terms of both $PSNR$ and $SSIM$. Second only to MCDnCNNs, our proposed model MCDnCNN for general noise is also better than other compared methods in terms of $PSNR$ from noise level 1\% to 15\%. and $SSIM $ from noise level 5\% to 15\%. .

\begin{figure}[t]
\centering
\quad
\subfigure[]{
\includegraphics[width=0.2\textwidth]{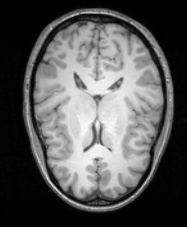}
}
\subfigure[]{
\includegraphics[width=0.2\textwidth]{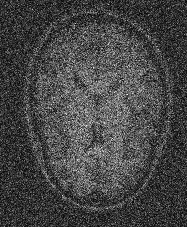}
}
\subfigure[]{
\includegraphics[width=0.2\textwidth]{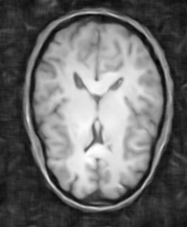}
}
\subfigure[]{
\includegraphics[width=0.2\textwidth]{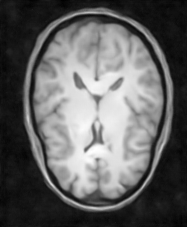}
}

\subfigure[]{
\includegraphics[width=0.2\textwidth]{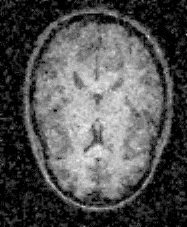}
}
\subfigure[]{
\includegraphics[width=0.2\textwidth]{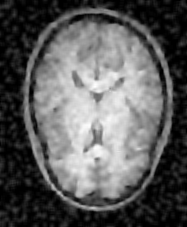}
}
\subfigure[]{
\includegraphics[width=0.2\textwidth]{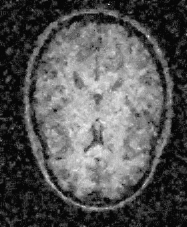}
}
\subfigure[]{
\includegraphics[width=0.2\textwidth]{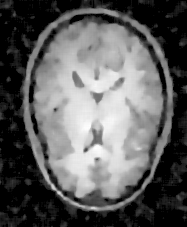}
}

\caption{\label{fig:example1} One denoising example with a noise-free image (a), the noisy image (b), denoised image from $MCDnCNNg$ (c), $MCDnCNNs$ (d), $Coupe-Block$ (e), $ODCT3D$ (f), $WSM$ (g) and $PRI-NLM3D$ (h). from dataset 1 }
\end{figure}

Figure \ref{fig:example1} shows an example of denoising result from different methods on data with 15\% noise. Visually, dedicated 15\% noise-specific model MCDnCNNs gives the most clean result. All other methods did not remove noise in the background sufficiently. In the brain foreground, both MCDnCNNs and MCDnCNNg achieve visually good result. All other methods removed noise but also lost details of the image.

\begin{table}
  \centering
\begin{tabular}{c c c c c c c c c}

  \hline			
  Method & $1\%$ & $3\%$  & $5\%$  & $7\%$  & $9\%$ & $11\%$ & $13\%$ & $15\%$ \\
  \hline			

  $Coupe-Block$ & 43.00&37.49&34.70&32.83&31.36&30.08&28.99&27.95\\
  $WSM$ &43.16&37.80&34.95&33.01&31.45&30.10&28.93&27.82\\  
  $ODCT3D$ &43.29&38.47&35.69&33.78&32.34&31.13&30.08&29.16\\
  $PRI-NLM3D$ &44.30&38.50&35.67&33.74&32.37&31.14&30.16&29.24\\
  $MCDnCNNg$ & 44.24 &39.38 &37.12 &35.40 &33.86 &32.54 &31.10 &29.96 \\
  $MCDnCNNs$ & \textbf{46.32}&\textbf{40.47}&\textbf{37.82}&\textbf{36.20}&\textbf{34.71}&\textbf{33.56}&\textbf{32.57}&\textbf{31.62} \\  
  \hline  
\end{tabular}
\caption{PSRN measure of different methods on different noise level from IXI-Hammersmith dataset}
\label{table:PSRN}
\end{table}

\begin{table}
  \centering
\begin{tabular}{c c c c c c c c c}

  \hline			
  Method & $1\%$ & $3\%$  & $5\%$  & $7\%$  & $9\%$ & $11\%$ & $13\%$ & $15\%$ \\
  \hline			

  $Coupe-Block$ & 0.98&0.94&0.90&0.85&0.81&0.77&0.74&0.70 \\
  $WSM$ & 0.98&0.94&0.90&0.85&0.81&0.77&0.73&0.69\\  
  $ODCT3D$ &0.97&0.97&0.93&0.89&0.86&0.83&0.80&0.77 \\
  $PRI-NLM3D$ & 0.99&0.96&0.92&0.89&0.86&0.83&0.89&0.78 \\
  $MCDnCNNg$ & 0.97&0.95&0.93&0.90&0.87&0.84&0.81&0.79 \\
  $MCDnCNNs$ & \textbf{0.99}&\textbf{0.96}&\textbf{0.93}&\textbf{0.92}&\textbf{0.89}&\textbf{0.87}&\textbf{0.85}&\textbf{0.83}\\
  \hline  
\end{tabular}
\caption{SSIM measure of different methods on different noise level from IXI-Hammersmith dataset  }
\label{table:SSIM}
\end{table}

\subsection{Results from IXI-Guys dataset}

\begin{figure}[t]
\centering
\quad
\subfigure[]{
\includegraphics[width=0.35\textheight]{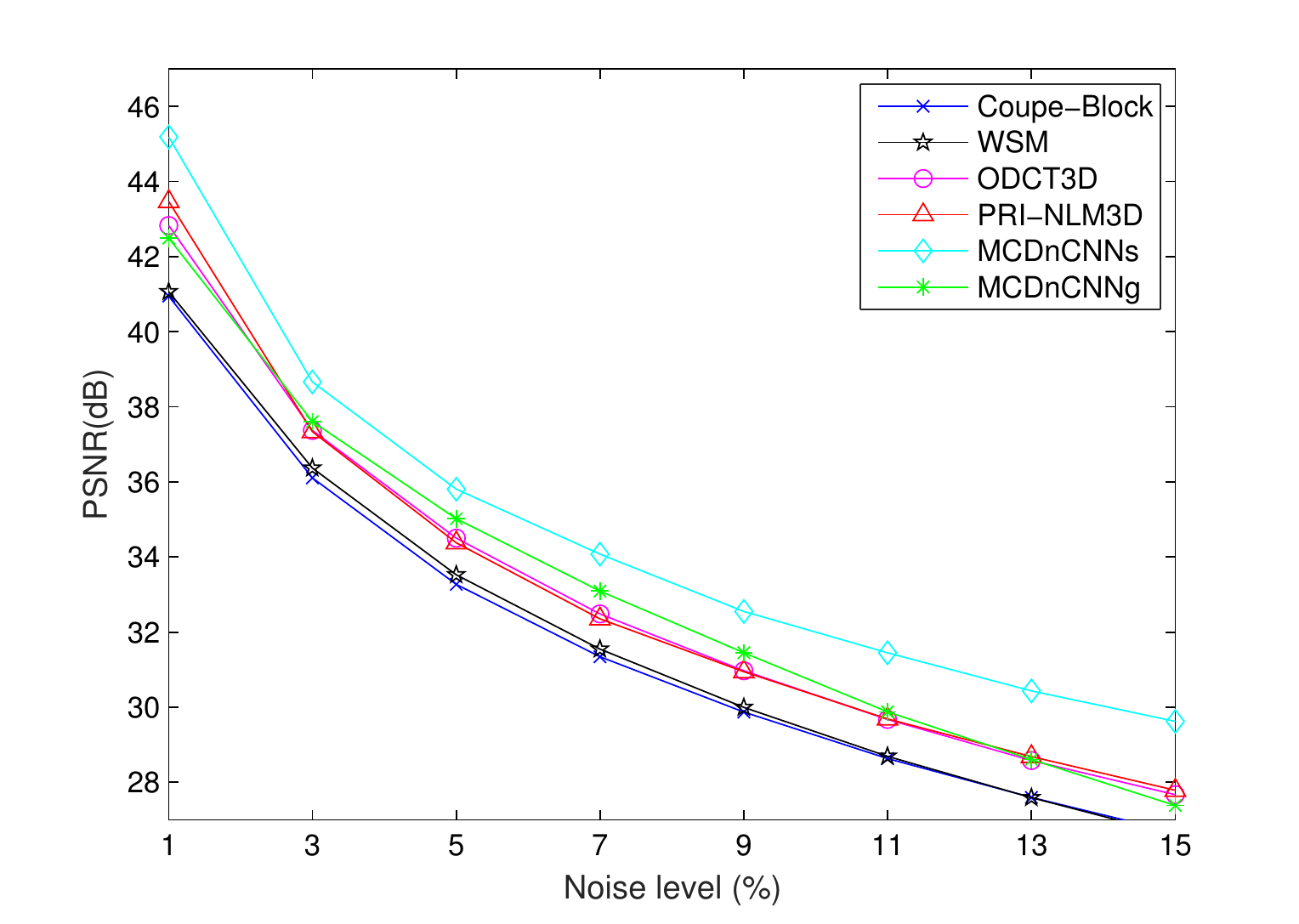}
}
\subfigure[]{
\includegraphics[width=0.35\textheight]{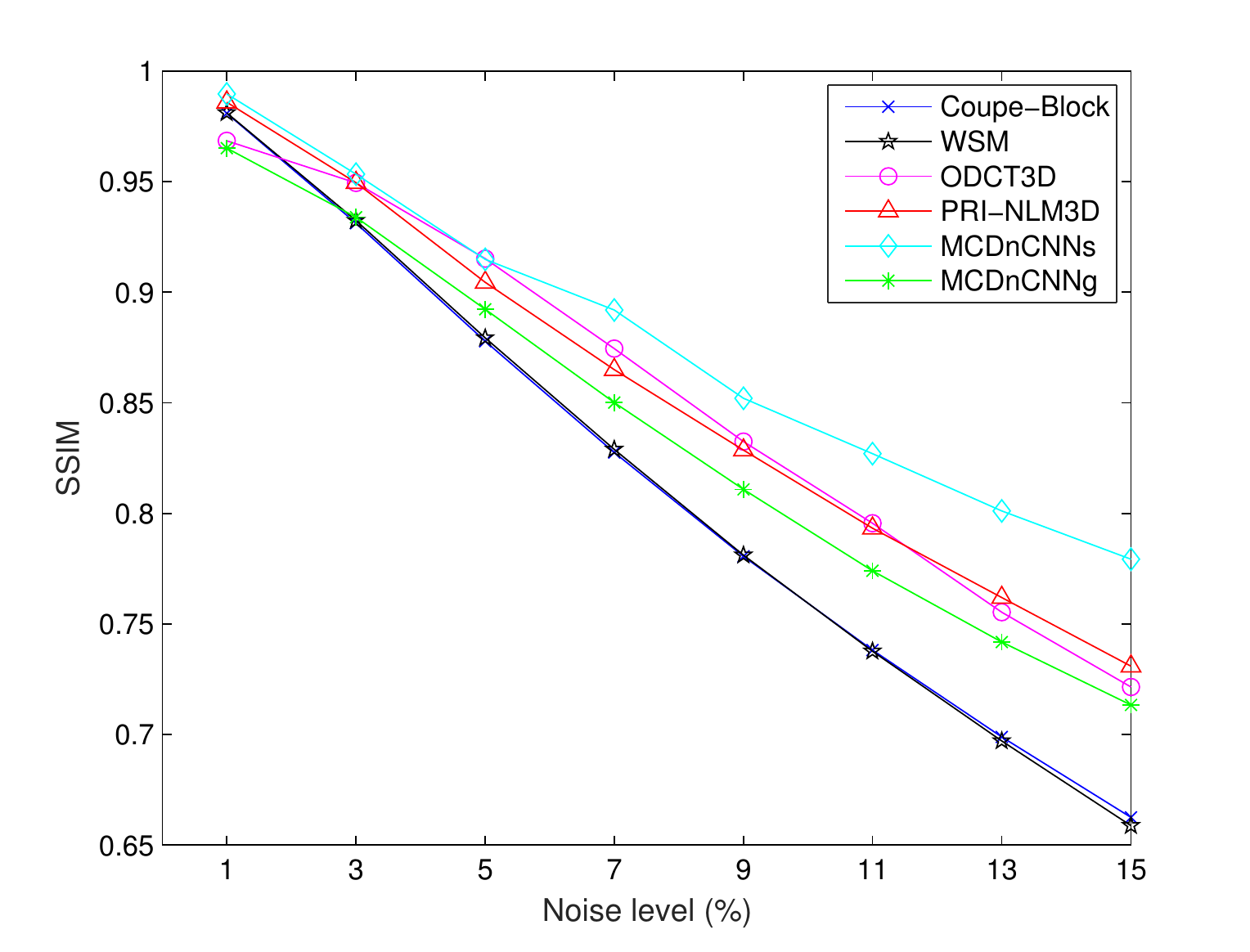}
}
\caption{\label{fig:measures_guys} De-nosie measures of PSRN (a) and SSIM (b) from different methods with different noise levels from Guys dataset (T1 1.5 T)}
\end{figure}

\begin{figure}[t]
\centering
\quad
\subfigure[]{
\includegraphics[width=0.2\textwidth]{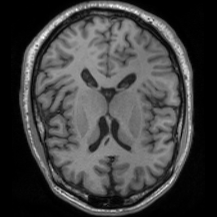}
}
\subfigure[]{
\includegraphics[width=0.2\textwidth]{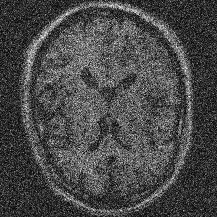}
}
\subfigure[]{
\includegraphics[width=0.2\textwidth]{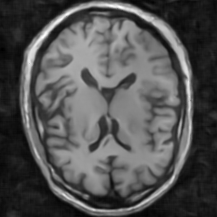}
}
\subfigure[]{
\includegraphics[width=0.2\textwidth]{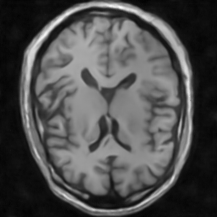}
}

\subfigure[]{
\includegraphics[width=0.2\textwidth]{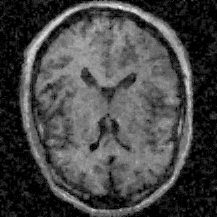}
}
\subfigure[]{
\includegraphics[width=0.2\textwidth]{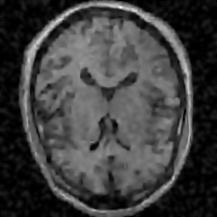}
}
\subfigure[]{
\includegraphics[width=0.2\textwidth]{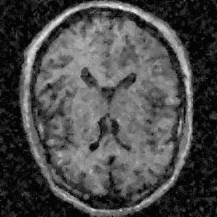}
}
\subfigure[]{
\includegraphics[width=0.2\textwidth]{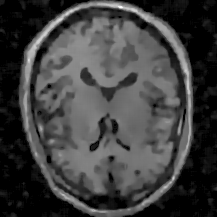}
}

\caption{\label{fig:guys} One denoising example with a noise-free image (a), the noisy image (b), denoised image from $MCDnCNNg$ (c), $MCDnCNNs$ (d), $Coupe-Block$ (e), $ODCT3D$ (f), $WSM$ (g) and $PRI-NLM3D$ (h). from Guys dataset }
\end{figure}

\begin{figure}[t]
\centering
\quad
\subfigure[]{
\includegraphics[width=0.35\textheight]{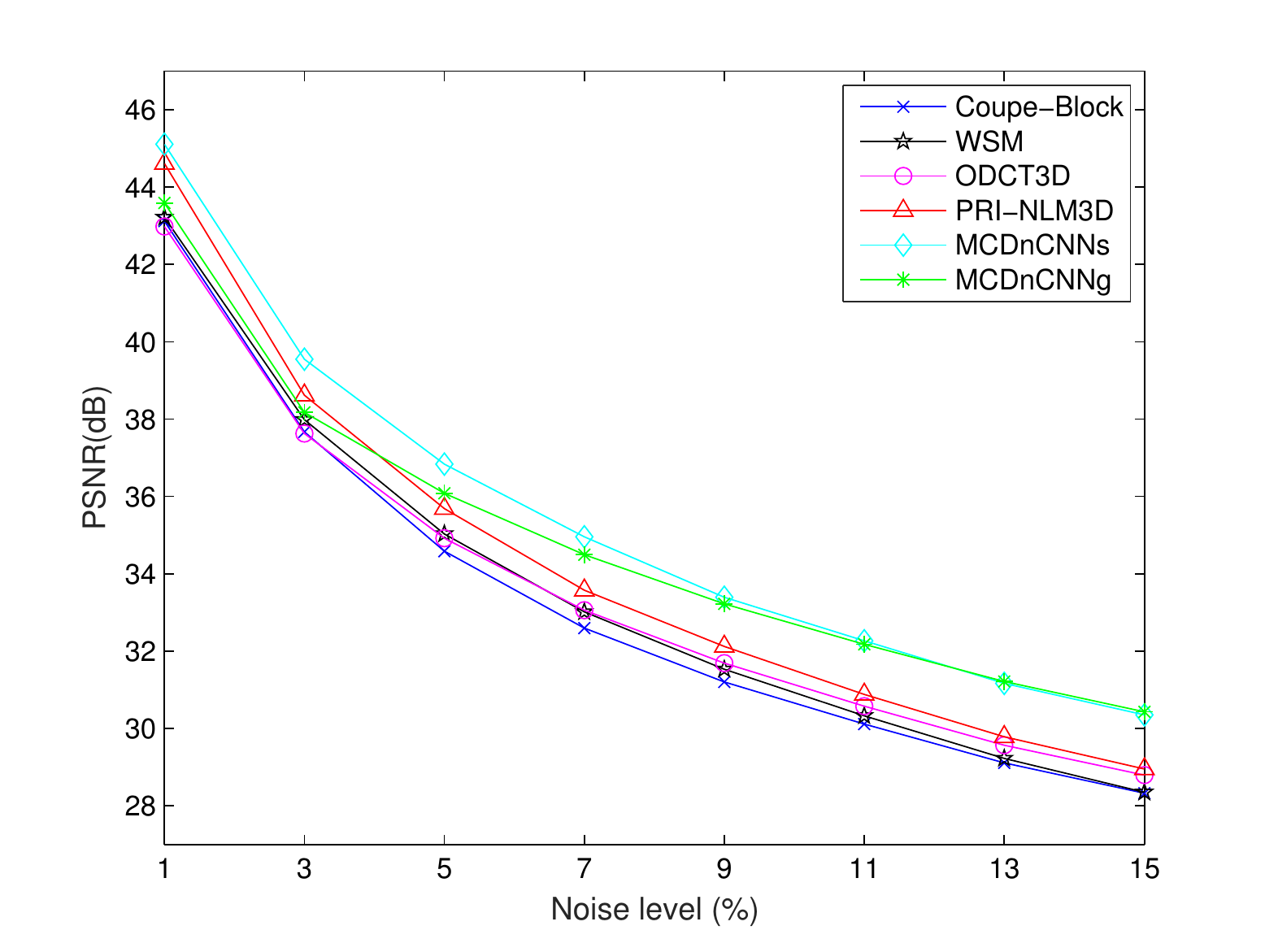}
}
\subfigure[]{
\includegraphics[width=0.35\textheight]{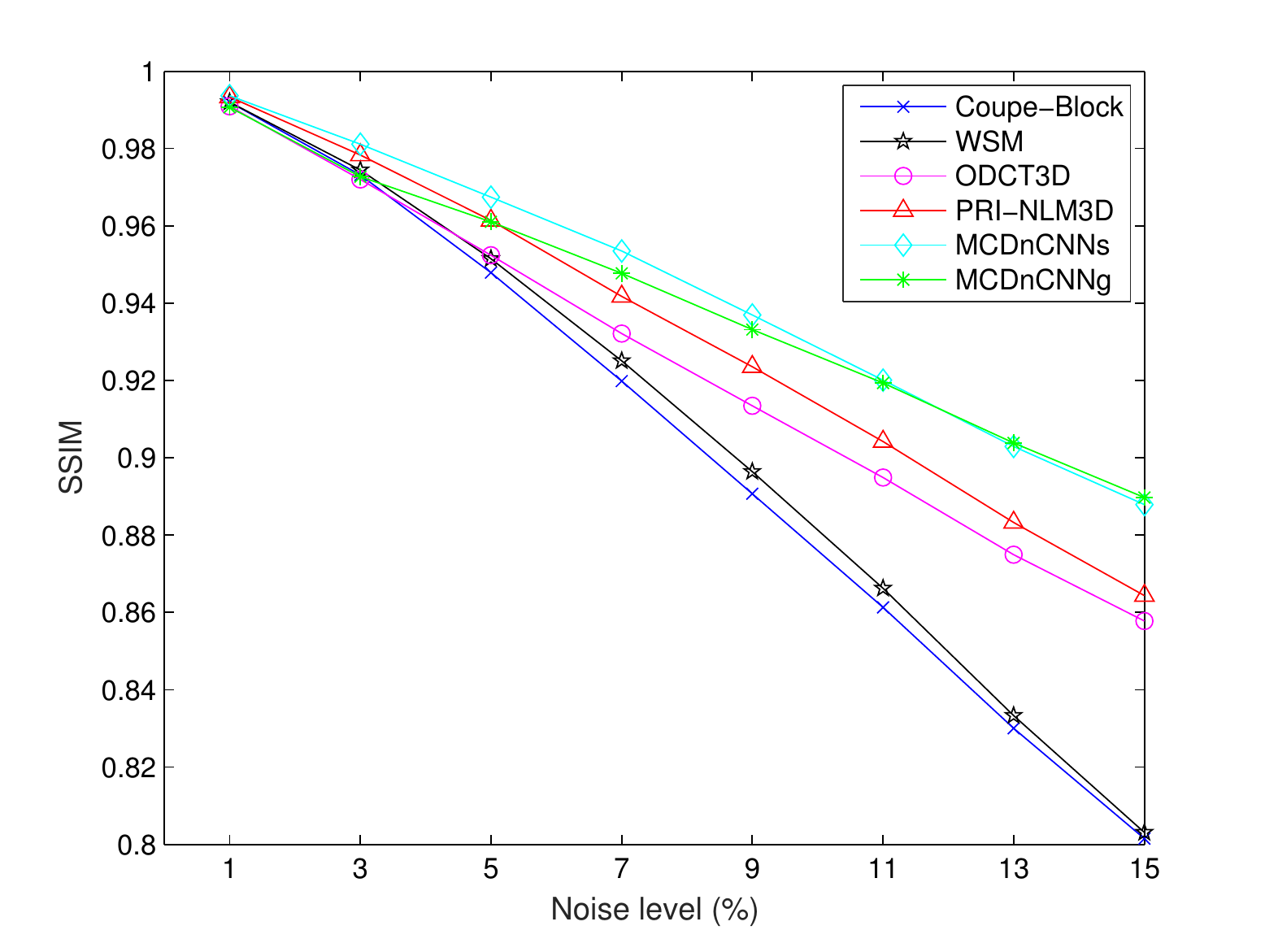}
}
\caption{\label{fig:measures_brainweb} De-nosie measures of PSRN (a) and SSIM (b) from different methods with different noise levels from Brainweb dataset}
\end{figure}

\begin{figure}[t]
\centering
\quad
\subfigure[]{
\includegraphics[width=0.2\textwidth]{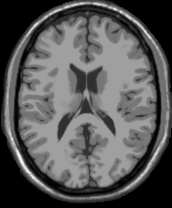}
}
\subfigure[]{
\includegraphics[width=0.2\textwidth]{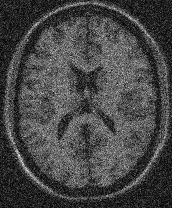}
}
\subfigure[]{
\includegraphics[width=0.2\textwidth]{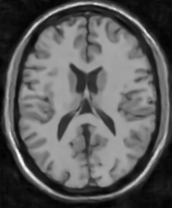}
}
\subfigure[]{
\includegraphics[width=0.2\textwidth]{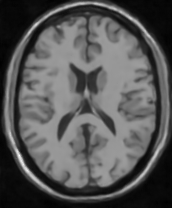}
}

\subfigure[]{
\includegraphics[width=0.2\textwidth]{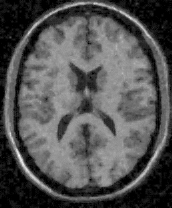}
}
\subfigure[]{
\includegraphics[width=0.2\textwidth]{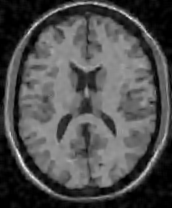}
}
\subfigure[]{
\includegraphics[width=0.2\textwidth]{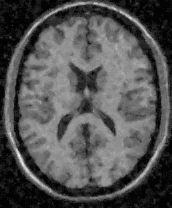}
}
\subfigure[]{
\includegraphics[width=0.2\textwidth]{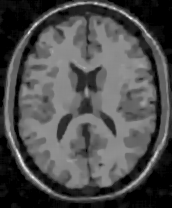}
}

\caption{\label{fig:brainweb} One denoising example with a noise-free image (a), the noisy image (b), denoised image from $MCDnCNNg$ (c), $MCDnCNNs$ (d), $Coupe-Block$ (e), $ODCT3D$ (f), $WSM$ (g) and $PRI-NLM3D$ (h). from Brainweb dataset}
\end{figure}

Figure \ref{fig:measures_guys} summarizes the PSNR and SSIM with different methods from Guys dataset. Although we did not train on this dataset, the results shows that our two models still achieved a good performance. The noise-specific model MCDnCNNs shows most robust performance compared to all other methods in terms of $PSNR$ and $SSIM$. Next to MCDnCNNs, our proposed model MCDnCNNg for general noise is also significantly better than other compared methods in terms of $PSNR$ from noise levels 3\% to 13\%. However in terms of $SSIM$, from noise levels 3\% to 15\%, ODCT3D and PRI\_NLM3D opposed our general model. It should be noted that both MCDnCNNg and MCDnCNNs are not retrained on this dataset.

Figure \ref{fig:guys} shows an example of denoising result from different methods on data with 15\% noise from Guys dataset. We can see that the effect is similar to that from the IXI-Hammersmith dataset.

\subsection{Results from Brainweb dataset}

Figure \ref{fig:measures_brainweb} summarizes the PSNR and SSIM in Brainweb dataset from different methods. In analogous to the finding shown above, the noise-specific model MCDnCNN (MCDnCNNs) significantly outperforms all other method in terms of $PSNR$ and $SSIM$. Second only to MCDnCNNs, our proposed model MCDnCNN for general noise (MCDnCNNg) is also significantly better than other compared methods in terms of $PSNR$ and $SSIM$ from noise levels 5\% to 15\%.

%
%
%
%

Figure \ref{fig:brainweb} shows an example of denoising result from different methods on data with 15\% noise from Brainweb dataset. We can see that the effect is similar to that from the IXI-Hammersmith dataset.

\section{Discussion and Conclusion}

In this paper, we present a novel approach to denoise robustly MR images based on the deep convolutional feed forward neural \cite{Zhang2017a}. This net is multi-channel-based combined with residual learning. Our proposed noise-specific feed forward denoising model outperforms all other methods in term of the peak signal to noise ratio and the global of structure similarity index in all our three test datasets. Moreover, our general noise-applicable model is also better than the other compared methods in two datasets. The reason why in one dataset the general model achieves not qualified performance is probably because the model needs to be retrained. In summary, testing with datasets including IXI-Hammersmith, IXI-guys and Brainweb data, our general model shows good general applicability and compare favorably to other conventional methods.

The advantage of our proposed method is that the new training scheme is within an unified framework. Our general denosing model is trained with the noisy images from a wide range of noise levels. To test a given image, this model can be applied without assuming or estimating its noise level in advance. Although the performance of our general noise-level model is slightly worse than our noise-level specific model (each model was trained with images with a specific noise level). The general model however can be applied for MR images directly out-of-box in real practice. It should be noted that all other compared methods need parameter-tuning for a specific noise level. Before applying the most robust denosing methods, the noise level on a specific image needs to be estimated and this estimation may affect denosing effect. To improve the performance, the general model can be further trained and tuned in the corresponding dataset.

In this paper, we demonstrated that our denoising model trained from one MR dataset with 3T can be directly applied on other two datasets with a different magnetic field strength and the performance of our noise-specific deep-learning model outperforms that from other methods in the other two datasets. It should be noted that the Brainweb dataset used here is simulated noise-free data and results from this dataset is most convincing. This validates that the general applicability of the deep-learning based denoise model.

In this paper, a deep convolutional neural network was proposed for image denoising. The max-pooling is not used to keep output the same size to that of input image. In the future, we are going to explore other possibilities such as deep convolutional generative adversarial networks(DCGAN)\cite{Denton2015,Goodfellow2014,Halbritter2017} and u-net. In u-net, although max-pooling is used, at the same network one upsampling scheme is applied to restore the size of the output image.

In our future work, instead of making use of a multi-channel network, we would investigate the effect of feeding the whole 3D volume to a complete 3D convolutional neural network. We would also like to explore the possibilities of extending this proposed method to remove the image noise in lung, cardiac or abdominal MRI studies.

\bibliographystyle{spiejour}   

\bibliography{./medlinestrings,./report}

\listoffigures
\listoftables

\end{document}